# The Nonparametric Metadata Dependent Relational Model

Dae Il Kim                                                                                       daeil@cs.brown.edu
Michael C. Hughes                                                                             mhughes@cs.brown.edu
Erik B. Sudderth                                                                             sudderth@cs.brown.edu

Brown University, Department of Computer Science, Providence, RI 02906 USA

## Abstract

We introduce the nonparametric metadata dependent relational (NMDR) model, a Bayesian nonparametric stochastic block model for network data. The NMDR allows the entities associated with each node to have mixed membership in an unbounded collection of latent communities. Learned regression models allow these memberships to depend on, and be predicted from, arbitrary node metadata. We develop efficient MCMC algorithms for learning NMDR models from partially observed node relationships. Retrospective MCMC methods allow our sampler to work directly with the infinite stick-breaking representation of the NMDR, avoiding the need for finite truncations. Our results demonstrate recovery of useful latent communities from real-world social and ecological networks, and the usefulness of metadata in link prediction tasks.

## 1. Introduction

A recent explosion in the magnitude and complexity of relational datasets motivates algorithms that can discover meaningful latent structure within complex, observed networks. Blocks of nodes with similar behavior are called *communities*, and could represent predators with common prey, proteins that regulate similar functions, or individuals with common interests.

Wang & Wong (1987) proposed the *latent stochastic blockmodel* (LSB) as a generalization of mixture models to relational datasets, in which each node is assigned to one of some finite set of communities. The *infinite relational model* (IRM) (Kemp et al., 2006) used a *Chinese restaurant process* (CRP) prior on partitions to group nodes into an unbounded set of communities, but still allocates each node to a single community. Airoldi et al. (2008) instead propose a *mixed membership stochastic block* (MMSB) model, where each node has a distribution over a finite set of latent communities, and chooses potentially distinct communities to generate each observed relationship. Parametric model selection methods are needed to choose the number of latent MMSB blocks. The *hierarchical Dirichlet process* (Teh et al., 2006) provides one approach to nonparametric mixed membership modeling, but it has not yet been applied to relational data.

Miller et al. (2009) describe a *nonparametric latent feature model* (NLFM), which uses an *Indian buffet process* (IBP, (Griffiths & Ghahramani, 2005)) prior to associate each node with a subset of an unbounded collection of latent features. Each instantiated feature then contributes to a generalized linear model of relationship probabilities. Nodal metadata can be incorporated into their edge likelihood model, but does not directly influence the generation of latent features. While the NLFM leads to useful link prediction algorithms, experiments suggested that the recovered features were difficult to interpret and provided little qualitative understanding of real networks. Choi et al. (2011) similarly incorporate metadata into their edge likelihoods, but within the context of a parametric LSB. Their focus is on estimating appropriate confidence sets for the inferred latent structure.

Some stochastic block models have modeled hierarchical structure within the latent communities, including the *Mondrian process* (Roy & Teh, 2009) and the *multiscale community block model* (MCSB, Ho et al. (2011a)). The MCSB uses a nested CRP prior (Blei et al., 2010) to associate each node with a finite-depth chain of communities, organized in a tree of potentially unbounded degree. Other block models are designed for dynamic network data (Ho et al., 2011b; Ishiguro et al., 2010; Fu et al., 2009), but our focus here is on static networks.

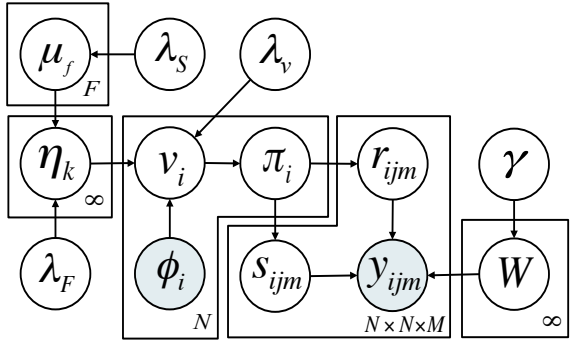

Figure 1. Directed graphical representation of an NMDR model for N nodes, and $M$ binary relations which potentially exist among each of the $N \times N$ node pairs. The community assignments $s_{ijm}$ (source) and $r_{ijm}$ (receiver) for relation $y_{ijm}$ depend on node-specific community frequencies $\{\pi_{:i}, \pi_{:j}\}$, which are deterministically mapped from Gaussian latent variables $\{v_{:i}, v_{:j}\}$ via a logistic function. The mean of $v_{ki}$ is $\eta_k^T \phi_{:i}$, allowing learned dependence on node metadata $\phi_{:i}$. The unbounded community interaction matrix $W$ is marginalized via a conjugate beta prior.

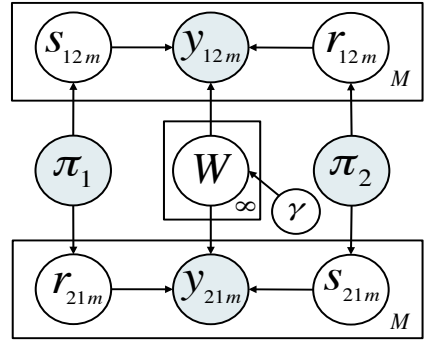

Figure 2. Directed graph for the generation of relations $y_{12m}, y_{21m}$ given $\pi_{:1}, \pi_{:2}$. Note that the first subscript in source variable $s_{12m}$ indicates which mixed memberships ($\pi_{:1}$) generate it, while the second subscript in receiver variable $r_{12m}$ indicates its generating memberships ($\pi_{:2}$).

The *nonparametric metadata dependent relational* (NMDR) model proposed in this paper extends prior stochastic blockmodels in two key ways: each node has mixed membership in an unbounded set of communities, and node-specific metadata directly influences the generation of these latent community distributions. In contrast to the simpler likelihood-based way the NLFM uses metadata, our approach leads to intuitive, interpretable communities which also provide accurate link predictions. The NMDR incorporates metadata via a logistic stick-breaking process (Ren et al., 2011), analogously to the *doubly correlated nonparametric topic* (DCNT) model (Kim & Sudderth, 2011). Unlike the DCNT, which employed a finite truncation for approximate MCMC inference, we use retrospective MCMC methods (Papaspiliopoulos & Roberts, 2008) to develop MCMC learning algorithms for the true, infinite NMDR model. We also develop a likelihood model appropriate for multi-relational networks. Our experiments show recovery of qualitatively interesting structure, and quantitatively accurate link prediction, for social and ecological networks.

## 2. Nonparametric Relational Modeling

The NMDR is a Bayesian nonparametric hierarchical extension of the MMSB. We provide an overview of the model focusing on our two major innovations, "upstream" incorporation of metadata (Mimno & McCallum, 2008) and a stick-breaking representation of the infinite mixed-membership vector.

### 2.1. Node Metadata

For $N$ nodes, we let $\phi_{:i} \in \mathbb{R}^F$ denote a feature vector[1] that captures the metadata associated with node $i$, and $\phi \in \mathbb{R}^{F \times N}$ a matrix of corpus metadata. When no metadata is available, we set $\phi_{:i} = 1$ to allow learning of the mean frequency of each community.

For every community $k$ in our unbounded set, we let $\eta_{fk} \in \mathbb{R}$ denote an associated significance weight for feature $f$ in community $k$, and $\eta_{:k} \in \mathbb{R}^F$ a vector of these weights. A Gaussian prior $\eta_{:k} \sim N(\mu, \Lambda^{-1})$ regularizes the topic weights, where $\mu \in \mathbb{R}^F$ is a vector of mean feature responses, and $\Lambda = \lambda_F I$ is a diagonal precision matrix. As in standard hierarchical Bayesian regression models (Gelman et al., 2004), we assign conjugate priors $\mu_f \sim N(0, \lambda_S^{-1})$, $\lambda_F \sim \text{Gam}(a_F, b_F)$, and $\lambda_S \sim \text{Gam}(a_S, b_S)$.

Given $\eta$ and $\phi$, the node-specific *score* for community $k$ is sampled as $v_{ki} \sim N(\eta_{:k}^T \phi_{:i}, \lambda_V^{-1})$. Slightly abusing notation, we can compactly write this transformation as $v_{:i} \sim N(\eta^T \phi_{:i}, L^{-1})$, where $v_{:i}$ is the infinite sequence of scores for node $i$ and $L = \lambda_V I$ is an "infinite" diagonal precision matrix. Note that the distribution of $v_{ki}$ depends only on the first $k$ entries of $\eta^T \phi_{:i}$, not the infinite tail of scores for subsequent topics; this makes our retrospective MCMC sampler tractable. The node-specific community scores, which may be any real numbers, are next converted to valid mixed-membership probability distributions.

### 2.2. A Logistic Stick-Breaking Transformation

To allow tractable learning of an unbounded set of communities, we employ a stick-breaking repre-sentation of the community distributions. Stick-breaking constructions are widely used in applications of Bayesian nonparametric models, and lead to convenient sampling algorithms (Ishwaran & James, 2001). Let $\pi_{ki}$ denote the probability that node $i$ chooses community $k$, where $\sum_{k=1}^{\infty} \pi_{ki} = 1$. The NMDR constructs these probabilities as follows:

$$\pi_{ki} = \psi(v_{ki}) \prod_{\ell=1}^{k-1} \psi(-v_{\ell i}) \quad (1)$$

$$\psi(v_{ki}) = \frac{1}{1+\exp(-v_{ki})} \quad (2)$$

---
[1] For any matrix $\eta$, we let $\eta_{:k}$ denote a column vector indexed by $k$, and $\eta_{f:}$ a row vector indexed by f.

Here, $0 < \psi(v_{ki}) < 1$ is the classic logistic function, which satisfies $\psi(-v_{\ell i}) = 1 - \psi(v_{\ell i})$. This same transformation is part of the so-called *logistic stick-breaking process* (Ren et al., 2011). However, they employ a very different prior distribution for $v_{ki}$ and are motivated by different applications. Our usage of this logistic stick-breaking transformation is inspired by the prior on topic distributions underlying the DCNT model (Kim & Sudderth, 2011).

### 2.3. Generating Relational Edges

The generative process described here applies to multi-relational, directed binary graphs. Generalizations to undirected graphs, or non-binary relations, can be accommodated by slight modifications. Given the mixed-membership distributions for nodes $i$ and $j$, $\pi_{:i}$ and $\pi_{:j}$, we sample a pair of community indicator variables for each directed interaction $y_{ijm}$. For relation $m$, source indicator $s_{ijm} \sim \text{Mult}(\pi_{:i})$ and receiver indicator $r_{ijm} \sim \text{Mult}(\pi_{:j})$. See Fig. 2.

Once the source and receiver communities have been chosen, binary edge $y_{ijm}$ is chosen from a corresponding Bernoulli distribution, $y_{ijm} \sim \text{Ber}(s_{ijm}^T W_{::m} r_{ijm})$. For each of the $M$ types of relations, we place a conjugate beta prior on the entries of the infinite block relation matrix $W_{::m}$, so that $W_{k\ell m} \sim \text{Beta}(\gamma_a, \gamma_b)$ for some fixed hyperparameters $\gamma_a, \gamma_b$.

Because the number of latent communities is unbounded, there are infinitely many relationship probabilities $W_{k\ell m}$, $M$ for each community pair. During learning, our inference algorithm marginalizes the entries of $W$. Resampling of community indicators $(s_{ijm}, r_{ijm})$, given fixed scores $v_{:i}, v_{:j}$, remains tractable. Our retrospective MCMC algorithm would remain simple for other likelihood functions with conjugate priors, for example Poisson for count relations, or Gaussian for real-valued relations.

In summary, the NMDR generative model is as follows:

1. Sample global parameters:
   (a) Draw $\lambda_S, \lambda_F, \lambda_V$ from gamma priors
   (b) Draw $\mu \sim N(0, \lambda_S^{-1} I_F)$
   (c) For each community $k$, $\eta_{:k} \sim N(\mu, \lambda_F^{-1} I_F)$

2. For each node $i = 1, 2, \ldots, N$:
   (a) Draw $v_{:i} \sim N(\eta^T \phi_{:i}, \lambda_V^{-1} I)$
   (b) Let $\pi_{ki} = \psi(v_{ki}) \prod_{\ell=1}^{k-1} \psi(-v_{\ell i})$

3. For each relation $m = 1, 2, \ldots, M$:
   (a) For each community pair $k, \ell$:
       i. Draw $W_{k\ell m} \sim \text{Beta}(\gamma_a, \gamma_b)$
   (a) For each potential edge $\{i, j\} \in N \times N$:
       i. Draw $s_{ijm} \sim \text{Mult}(\pi_{:i})$
       ii. Draw $r_{ijm} \sim \text{Mult}(\pi_{:j})$
       iii. Draw $y_{ijm} \sim \text{Ber}(s_{ijm}^T W_{::m} r_{ijm})$

## 3. Learning via MCMC

In applying the NMDR model to real-world networks, we observe some edges $y$ linking a set of nodes and wish to recover the block community structure, parameterized by $\eta, v, s, r, W$. In general, potential relations may be observed to be present ($y_{ijm} = 1$), observed to be absent ($y_{ijm} = 0$), or be unobserved and unknown ($y_{ijm} = ?$). Assuming unobserved relations are missing at random, as in our experiments, we assign those links uninformative likelihoods during learning.

Given observed relationships for some or all node pairs, we employ a *Markov chain Monte Carlo* (MCMC) method to draw samples from the hidden variables' posterior distribution. As a primary contribution, our MCMC algorithm performs retrospective moves which allow us to dynamically vary the number of instantiated latent communities.

### 3.1. Retrospective MCMC

Bayesian nonparametric models based on unconventional stick-breaking priors often use a finite, truncated approximation to the true infinite model. While this approach can be effective (Ishwaran & James, 2001), selection of an appropriate truncation level $K$ is challenging. When $K$ is conservatively large, substantial computational resources can be expended resampling "wasted" variables, and model interpretability often suffers. When $K$ is small, learning and inference are potentially biased, and the benefits which originally motivated the nonparametric approach are lost. To avoid these issues, we implement a dynamic truncation technique based on retrospective sampling (Papaspiliopoulos & Roberts, 2008) of latent community assignments $s$ and $r$.

Consider the resampling of a source indicator $s_{ijm}$, for relation $m$ from node $i$ to node $j$, given fixed values of all other indicators and variables. A similar approach can be used for resampling $r_{ijm}$, or for blocked resampling of $\{s_{ijm}, r_{ijm}\}$. Because we employ a conjugate beta prior, our sampler analytically marginalizes the relation parameters $W_{k\ell m}$, expressing the posterior in terms of various edge counts. Suppose that $r_{ijm} = \ell$. Excluding node pair $(i,j)$, let $A_{k\ell m}^{\backslash ij}$ denote the number of directed edges, for relation $m$, from nodes whose indicators associate that pairing to communities $(k, \ell)$. Similarly, let $B_{k\ell m}^{\backslash ij}$ denote the number of such node pairs which do not exhibit relation $m$.

Let $K$ denote the index of the largest community, in stick-breaking order, which currently has at least one assigned node. The retrospective sampler explicitly instantiates $v_{k:}$ and $\eta_{:k}$ for $k \leq K$. Computing $\pi_{ki}$ based on these variables, as in Eq. (1), we let

$$\rho_k \propto \pi_{ki} \left( \frac{(A_{k\ell m}^{\backslash ij} + \gamma_a)^{y_{ijm}} (B_{k\ell m}^{\backslash ij} + \gamma_b)^{1-y_{ijm}}}{A_{k\ell m}^{\backslash ij} + B_{k\ell m}^{\backslash ij} + \gamma_a + \gamma_b} \right) \quad (3)$$

for $k = 1, \ldots, K$, and

$$\rho_{K+1} \propto \left(1 - \sum_{k=1}^{K} \pi_{ki}\right) \left( \frac{\gamma_a^{y_{ijm}} \gamma_b^{1-y_{ijm}}}{\gamma_a + \gamma_b} \right). \quad (4)$$

The proportionality constant in Eqs. (3, 4) is selected so that $\rho$ is a properly normalized $(K+1)$-dimensional multinomial distribution. For $k \leq K$, $\rho_k$ is the posterior probability of selecting community $k$. $\rho_{K+1}$ is the aggregate posterior probability of the infinite "tail" of communities with indexes greater than $K$.

Algorithm 1 begins by sampling from $\text{Mult}(\rho)$ to determine whether $s_{ijm}$ should be assigned to one of the already instantiated communities, or some new community. If the sampled $s_{ijm} \leq K$, as is common after the first few sampling iterations, we simply choose that community. If not, we select a new community by simulating the logistic stick-breaking prior, since all potential new communities have indistinguishable marginal likelihoods. Such dynamic creation of variables is the key to retrospective samplers. Because our likelihood parameters $W_{k\ell m}$ have conjugate beta priors, we can exactly compute the posterior normalization constant, and the more complex Metropolis-Hastings proposals of Papaspiliopoulos & Roberts (2008) are unnecessary. A related approach has been used for inference in infinite depth nested CRP models (Blei et al., 2010).

**Algorithm 1** Retrospective MCMC resampling of source community $s_{ijm}$, given a $(K+1)$-dim. posterior distribution $\rho$ defined as in Eqs. (3, 4).

1: Draw $s_{ijm} \sim \text{Mult}(\rho)$
2: **if** $s_{ijm} = K + 1$ **then**
3:   Draw $\eta_{:s_{ijm}} \sim N(\mu, \Lambda^{-1})$
4:   Draw $v_{s_{ijm}i} \sim N(\eta_{:s_{ijm}}^T \phi_{:i}, \lambda_V^{-1})$
5:   Draw $\omega \sim \text{Ber}(\psi(v_{s_{ijm}i}))$
6:   **if** $\omega = 1$ **then**
7:     Exit and keep all instantiated variables
8:   **else**
9:     Increment $s_{ijm} \leftarrow s_{ijm} + 1$
10:    Goto line 3
11:  **end if**
12: **end if**

### 3.2. Conventional MCMC

Given $s$ and $r$, all but a finite subset of the variables in the infinite NMDR model are conditionally independent of the observed data. For most of these variables, our selection of conjugate priors allows closed form Gibbs sampling updates. For the node-specific community activation variables $v_{:i}$, which have a non-conjugate likelihood due to the logistic stick-breaking transformation, we instead use a Metropolis-Hastings independence sampler. As in (Kim & Sudderth, 2011), we find that repeated proposals from the prior are sufficient for adequate mixing. Please see the supplemental material for detailed resampling equations and the per-iteration cost of NMDR inference.

## 4. Results

We examine the NMDR's performance on two toy datasets as well as three real-world networks. To measure quantitative performance, we consider link prediction tasks and compare with publicly available implementations of MCMC algorithms for the MMSB (Chang, 2011) and IRM (Kemp et al., 2006). We evaluate via the area under the ROC curve (AUC). For unobserved edges, link probabilities are predicted from posterior samples via a straightforward Bayesian approach, detailed in the supplement. We also qualitatively examine learned communities, and find that the NMDR model captures a nuanced and useful mixed membership community structure.

### 4.1. Synthetic Data

We analyze two toy datasets, each with 80 nodes, generated by variants of stochastic block models. The first, `SynthSingle`, is a network where each node is assigned to exactly one of 5 blocks. Blocks have high within-block link probability and low between-block link probability, as shown in Fig. 3. Alternatively, the noisier `SynthMixed` network exhibits significant mixed membership among 4 blocks (see Fig. 3).

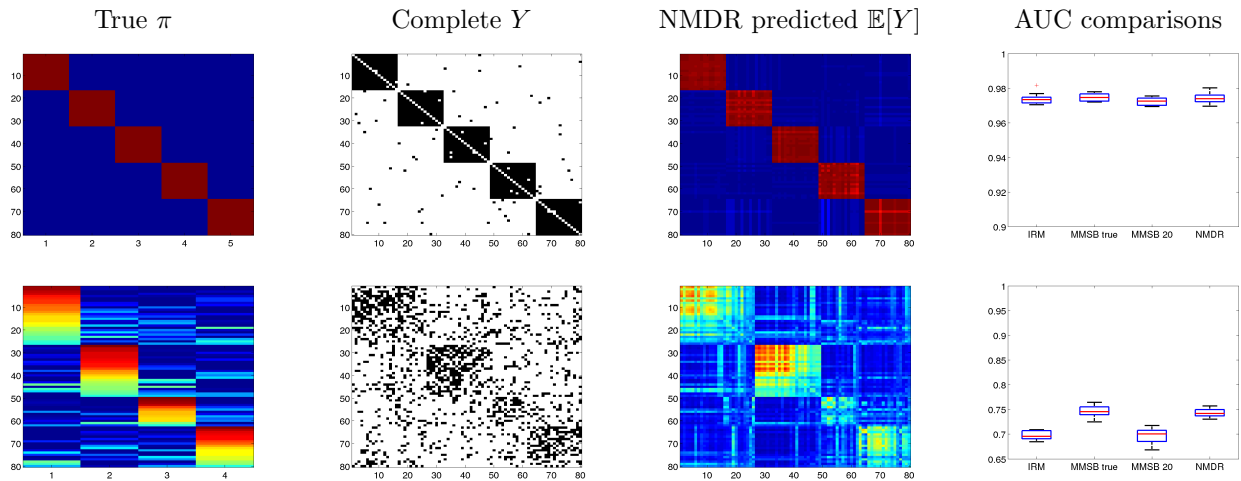

*Figure 3.* Single membership *(top)* and mixed membership *(bottom)* synthetic datasets. *Left to right:* True community memberships $\pi$. Complete relationship data $Y$. Relationship predictions from NMDR model learned from half of entries in $Y$. Box plots across 10 trials of link prediction AUC scores, based on most probable MCMC run for each model.

For each dataset, we simulate the edge removal process ten times, marking node pairs as unobserved at random with probability 0.5. We train NMDR, IRM, and MMSB models on the remaining edges. We run 3 MCMC chains for each model (6 for NMDR) for 6000 iterations, and use the best chain (in joint log probability) for prediction. We train two MMSB model variants, one with the true number of blocks, and one with $K = 20$. To evaluate, we report the spread in AUC scores across all 10 random masks in Fig. 3.

While the simple `SynthSingle` data matches the IRM's generative assumptions, the more flexible NMDR model nevertheless performs comparably to other methods. NMDR recovers the true block structure and has near-optimal prediction performance.

On `SynthMixed` data, we find that the NMDR model again has the highest performance (AUC $\approx$ 0.75), comparable to an MMSB based on the true number of blocks. The NMDR recovers an expected adjacency matrix $\mathbb{E}[Y]$ in close agreement with the true data, while the IRM suffers from not allowing mixed memberships. Given only a coarse upper bound on the true number of communities, MMSB performance also drops (AUC $\approx$ 0.70). This illustrates the benefits of our better regularized nonparametric model, and the ability of retrospective MCMC methods to automatically infer an appropriate set of communities, avoiding an expensive model selection process.

### 4.2. Sampson Monastery

Sampson's 1968 investigation of a monastery provides a benchmark dataset for qualitative network analysis (Sampson, 1968). Over a 12 month period, he observed interactions between 18 novice monks and surveyed feelings about their peers. Eight relations were recorded: like, dislike, influence, non-influence, esteem, disesteem, praise, and sanctions.

Sampson described two competing ideological factions. The "Young Turks", led by Gregory and John Bosco, consisted mostly of a new wave of brothers arriving during month 5 who questioned monastery practices. The "Loyal Opposition" faction, led by Peter, were mostly present from month 1 and favored the status quo. Sampson also noted an "Outcast" group, who were socially rejected by their peers, and "Interstitials," who oscillated between the dominant groups. We take Sampson's factions as a basis for judging the validity of our model's recovered block structure.

Our analysis differs in key aspects from the MMSB analysis in (Airoldi et al., 2008). First, we use data from all 8 surveys, rather than a curated single-relation summary. Second, we incorporate metadata found in Sampson's original thesis. This includes each monk's arrival and departure time, each monk's rank (1 of 4 categories), and binary judgments on sociability and maturity passed by monastery leaders.

Fig. 4 shows recovered mixed membership structure for each monk, obtained after 10,000 iterations of MCMC inference. Two of these blocks contain individuals from the "Young Turks" faction, with one of the blocks serving as the primary affiliation for all basic members and another through which leaders (Gregory and John Bosco) act on occasion. Another two blocks correspond to the "Loyalist" faction, with one community attracting all members and another assigned mostly to the hardened core of the group (e.g., Peter). Finally, the model allocates all three outcasts to a fifth "Outcast" block. An MDS visualization shows excellent correspondence with Sampson's primary factions as well as one interstitial node (Amand).

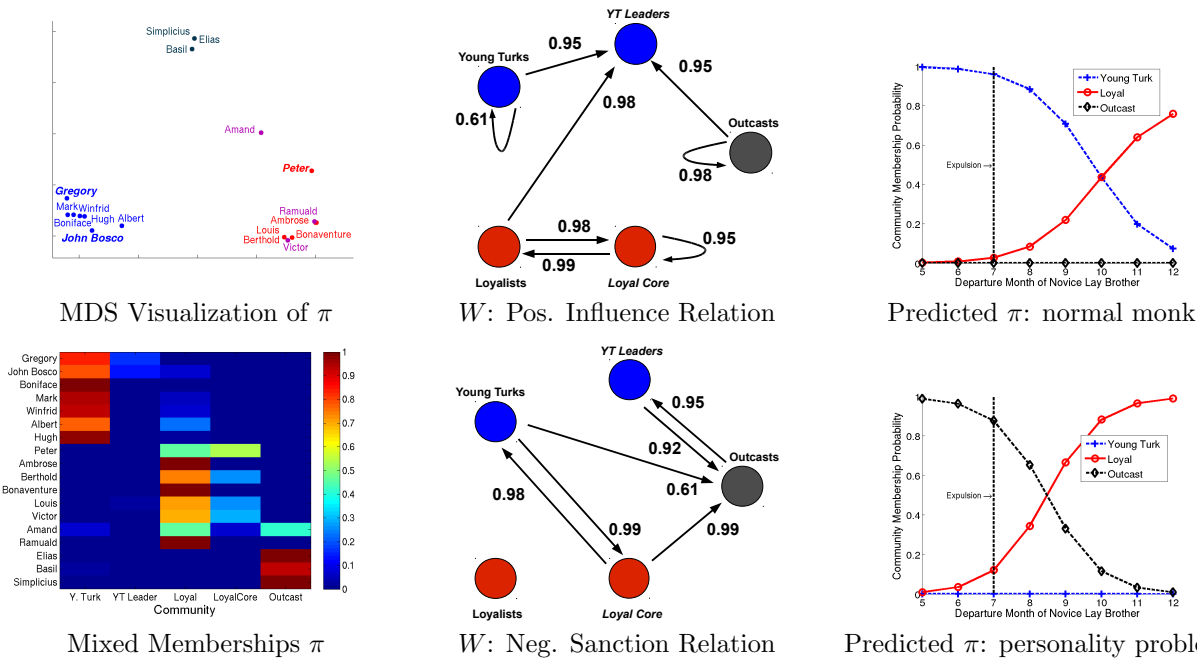

Figure 4. Recovered NMDR block structure for Sampson dataset, trained with metadata. *Top Left*: MDS visualizations of community memberships. Colors indicate Sampson's faction labels (black="Outcast", blue="Young Turk", red="Loyalist", purple="Interstitial"). *Bottom Left*: Estimated mixed membership probabilities for each monk. We manually label how communities intuitively align with Sampson's factions. *Center*: Learned link probabilities $W$ for influence (top) and sanction (bottom) relations. Links are drawn between communities $a, b$ if $W_{ab} > 0.1$. *Right*: Influence of metadata on community predictions. Changing key metadata (binary sociability flag, integer departure month) influences the learned prior on block membership. We build a $\phi$ for two novel monks with no observed links: one normal (top) and the other marked as a personality problem (bottom). Each plot shows that monk's $\pi$ for three primary communities as we vary the month departed covariate. We set $\eta$ to its final MCMC sample, and set $v$ to its MAP estimate.

To better understand why the model differentiates between leaders and followers, we visualize the learned $W$ for two key relations (influence and sanctions) in Fig. 4. Gregory and John Bosco were influential to monks from all factions (including some Loyalists), and together the pair alone received > 30% of all influence links. The model uses the "YT Leader" block to capture this difference. Among negative sanctions, Sampson's raw data finds most reprimands directed at outcasts, with the rest occuring between competing factions. Both realities are reflected by our model.

We find that the model learns intuitive relationships between block membership and the provided metadata, particularly departure times and personality judgments. Consider a novice lay brother for whom we have no observed relations. Intuitively, varying his "MonthDeparted" covariate should shift his affiliation. Fig. 4 shows our model predictions: departure near Gregory's expulsion implies "Young Turk" affiliation, while remaining longer implies "Loyalist" leanings. Similarly, observing personality problems implies "Outcast" status, unless he remains past expulsion.

### 4.3. Otago Harbour Food Web

The Otago Harbour dataset (Mouritsen et al., 2011) contains a single "who-eats-whom" binary relationship for 123 organisms from an intertidal mudflat ecosystem in New Zealand. In addition to predation links, the dataset contains metadata that broadly classifies each node as one of 21 possible organism types (e.g., annelids, birds), and assigns one of three mobility ratings (low, intermediate, high). A variety of organisms populate this food web including secondary predators (ducks, fish), primary predators (rock crabs), and autotrophs (seagrass). We explore whether unsupervised

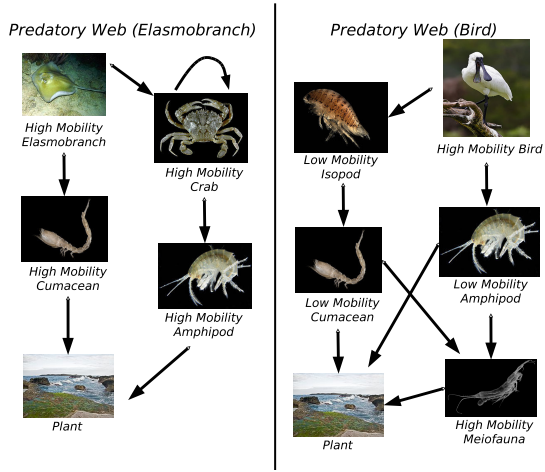

*Figure 5.* Food webs traced from two top predator archetypes, a *high mobility elasmobranch* (sharks, rays, and skates) and a *high mobility bird*. Given learned connections between metadata and predation relationships, links are drawn from each node to the most likely prey archetypes, continuing until the bottom of the web. The NMDR recovers biologically relevant structure from binary relations among 123 species. Images courtesy Wikipedia.

learning from metadata can contribute to knowledge discovery in complex, real-world networks.

We train an NMDR model on the entire graph, including metadata, and use our learned parameters to generate a food web relating organism archetypes. A node now represents a combination of organism and mobility types (e.g., high mobility bird, low mobility snail). As detailed in the supplementary material, edges represent a thresholded probability that organisms of these archetypes prey on one another.

By relating metadata archetypes, the graphs of Fig. 5 abstract the directly observed relationships among individual species. It could help scientists understand the canonical predator-prey relationships in this ecosystem. The top two prey choices often reflect meaningful biological relationships. For example, consider the self-loop for the high-mobility crab archetype. Carefully examining the raw data, every crab species cannibalistically preys on other crabs, including their own species. Furthermore, as expected the likelihood of being consumed by smaller organisms was near zero for top predators like birds and stingrays.

### 4.4. Lazega Lawyers Network

The Lazega lawyers dataset (Lazega, 2006) is a social network between partners and associates of several New England law firms. Collected from 1988-1991, it contains three directed binary relations encoding friendship, coworker, and advisory relationships among 71 lawyers. The dataset contains a rich set of metadata describing status, gender, office location, years employed, age, practice, and law school.

We focus on the quantitative task of link prediction, and compare our NMDR model (with and without metadata) to the IRM and MMSB.[2] Results are shown in Fig. 6. Even without metadata, the NMDR model performs much better than the IRM, demonstrating the importance of mixed memberships in modeling the complex relationships of real social networks. Compared to the MMSB, the NMDR without metadata performs similarly or slightly better. The small improvements might be attributed to the NMDR's ability to learn the number of communities dynamically via retrospective MCMC. Including metadata, the NMDR significantly outperform both the MMSB and IRM, showing the importance of incorporating relevant side information when it is available.

Finally, we show the benefit of using learned community memberships for visualization. Plotting the raw relationship graph, using the *GraphViz* (Ellson et al., 2001) force-atlas layout algorithm with equally weighted edges, results in a complex and noisy representation of the data (Fig. 6(a)). Instead, we can use the variational distance $D_{ij} = \frac{1}{2} \sum_{k=1}^{K} |\pi_{ki} - \pi_{kj}|$ between learned community membership distributions to measure node similarity in a more refined way. We assign node affinity weights of $1 - D_{ij}$, threshold these weights at $\frac{1}{2}$ to produce a sparse graph, and again apply the *GraphViz* force-atlas layout. This produces the far more intuitive and informative graph of Fig. 6(b).

## 5. Discussion

The NMDR model allows both discrete and continuous metadata to inform the community memberships of individual nodes. Retrospective MCMC methods allow data-driven determination of the number of latent communities, while avoiding expensive and potentially inaccurate truncations during learning. Our experiments suggest that the NMDR leads to competitive link prediction algorithms, which can further enhance accuracy by modeling metadata or multiple relations. The intuitive community structures recovered from real-world datasets are especially of interest, given interpretability problems reported with some prior nonparametric relational models.

---

[2]The MMSB was excluded from comparisons involving multiple relations, due to limitations of the MCMC inference code available in the `lda` R package (Chang, 2011).

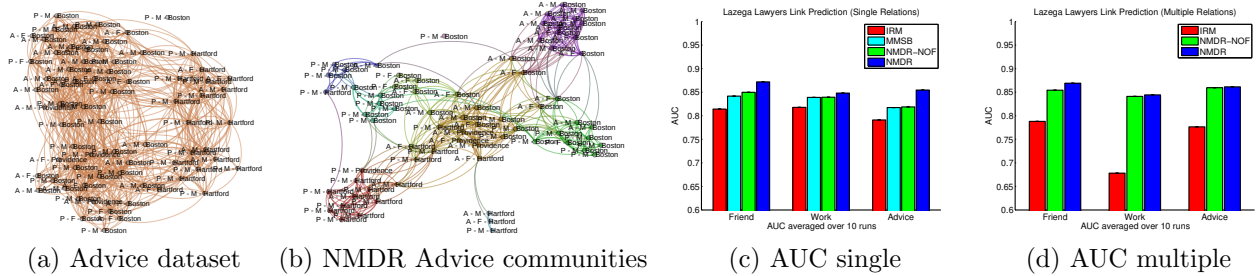

(a) Advice dataset  (b) NMDR Advice communities  (c) AUC single  (d) AUC multiple

*Figure 6.* Visualization and link prediction for the Lazega lawyers dataset. (a) Visualization of equally weighted edges for the "Advice" relation. Nodes are labeled by lawyer status (**P**artner or **A**ssociate), gender (**F**emale or **M**ale), and office location. (b) Visualization of weighted edges determined by the variational distance between learned community membership distributions. Nodes are colored according to their most associated community. (c,d) Mean AUC scores across 10 random masks for models of each individual relation. For each mask, half of the potential relationships are missing at random and must be predicted. For "Single" relation experiments (c), only data for the relation under consideration is available for training. For "Multiple" relation experiments (d), complete graphs for the other two relations are provided in addition to half of the potential target relations. AUC measures were calculated based on $\mathbb{E}[Y]$, marginalizing over latent variables. The NMDR is superior to the IRM even without metadata features (NMDR-NOF), and including metadata improves accuracy further. A Wilcoxin signed rank test confirms that the NMDR is significantly better than the IRM in all cases ($p < 0.05$). Without metadata, the NMDR-NOF performs comparably to the MMSB. With metadata, the full NMDR is significantly better ($p < 0.05$) than the MMSB for the Friend and Advice relations.

## Acknowledgments

We thank Prof. Edo Airoldi for helpful information regarding the Sampson monastery data. This research supported in part by IARPA under AFRL contract number FA8650-10-C-7059. D. Kim and M. C. Hughes supported in part by NSF Graduate Research Fellowships.